\documentclass{Interspeech}

\interspeechcameraready 

\title{A Practitioner's Guide to Building ASR Models for Low-Resource Languages: \\ A Case Study on Scottish Gaelic}
    
\author[affiliation={1}]{Ondřej}{Klejch}
\author[affiliation={2}]{William}{Lamb}
\author[affiliation={1}]{Peter}{Bell}

\affiliation{Centre for Speech Technology Research}{University of Edinburgh}{United Kingdom}
\affiliation{Celtic and Scottish Studies}{University of Edinburgh}{United Kingdom}
\email{\{o.klejch,w.lamb,peter.bell\}@ed.ac.uk}

\usepackage{tabularx}
\usepackage{blindtext}
\usepackage{tikz}
\newcommand{\tableref}[1]{{Table~\ref{#1}}}
\newcommand{\sectionref}[1]{{Section~\ref{#1}}}

\begin{document}

\maketitle

\begin{abstract}
An effective approach to the development of ASR systems for low-resource languages is to fine-tune an existing multilingual end-to-end model. When the original model has been trained on large quantities of data from many languages, fine-tuning can be effective with limited training data, even when the language in question was not present in the original training data. The fine-tuning approach has been encouraged by the availability of public-domain E2E models and is widely believed to lead to state-of-the-art results. This paper, however, challenges that belief.
We show that an approach combining hybrid HMMs with self-supervised models can yield substantially better performance with limited training data. This combination allows better utilisation of all available speech and text data through continued self-supervised pre-training and semi-supervised training. We benchmark our approach on Scottish Gaelic, achieving WER reductions of 32\% relative over our best fine-tuned Whisper model.

\end{abstract}

\section{Introduction}

Automatic speech recognition (ASR) has been democratised by several public-domain multi-lingual end-to-end (E2E) models.
For example, OpenAI's Whisper~\cite{radford2023robust} is a state-of-the-art multilingual model that supports around 100~languages.
Another example, Meta's MMS~\cite{pratap2024scaling}, supports more than 1000 languages.
However, this is only a fraction of all 7000 languages spoken world-wide.
Since these models were trained on millions of hours from many languages,
  they can be fine-tuned even for languages and domains unseen during their training.
Popular toolkits, such as HuggingFace Transformers~\cite{wolf2020transformers},
  allow laypeople to readily train ASR models for their particular language.%

On one hand, E2E models and toolkits are very accessible to beginners.
However, this accessibility comes at the cost of flexibility and finer control --
  these toolkits offer limited options for improving the performance of E2E models beyond simple hyperparameter tuning.
Furthermore,
  fine-tuning of E2E models -- which requires transcribed audio data -- might not be the optimal choice when having access only to a limited amounts of such data.
On the other hand, toolkits for traditional hybrid HMM systems,
  such as Kaldi~\cite{povey18_interspeech},
  have a much steeper learning curve. 
However, once grasped,
  they provide practitioners with finer control over every modelling decision,
  enabling them to train better models with limited training data.
They can also leverage unpaired text and audio more straightforwardly than E2E models.
Nevertheless, even if both types of systems have access to the same data, 
  hybrid models have been shown to work better even with 1000 hours of transcribed speech~\cite{rouhe2023principled}.

For example, hybrid models can more easily utilize language models trained on all available text corpora for the language in question.
These language models can be trained on vast amounts of text crawled from the internet
  and filtered by a language ID, in datasets like MADLAD-400~\cite{kudugunta2024madlad} or FineWeb~\cite{penedo2024fineweb}.
Hybrid models can also work with any language model (LM) type, ranging from n-gram LM, RNN LM~\cite{mikolov2010recurrent} to large Transformer LM~\cite{vaswani2017attention}.
Most recently, hybrid models have successfully leveraged large amounts of untranscribed speech by incorporating self-supervised pre-training~\cite{babu2022xlsr,chen2024towards}.
They can use large pre-trained self-supervised models (SSL models) as feature extractors similar to multilingual bottleneck features in the past~\cite{vesely2012language}.
Furthermore, these SSL models can be adapted to the target language with self-supervised continued pre-training~\cite{lee2022ctrl}
using only audio data. 

Hybrid models can also be improved with semi-supervised training~\cite{lamel2002,manohar2018semi},
  which uses a seed ASR model to produce pseudo-labels for untranscribed speech.
These pseudo-labels can then be used for standard supervised training.
Semi-supervised training can significantly improve a poor acoustic model if there is access to a good language model~\cite{wallington2021learning}:
  for example a language model trained on the aforementioned web-scale text corpora.
Furthermore, it has been shown that self-supervised pre-training and semi-supervised training are complementary~\cite{xu2021self} and can be used to improve the performance of ASR systems for low-resource languages~\cite{lam2023comparing}.
E2E models can also be improved with semi-supervised training~\cite{park2020improved},
  but since they cannot so easily leverage knowledge from text,
  the performance of semi-supervised training with E2E models is likely to be bounded by the quality of the seed model.
Still, it is possible to train a good hybrid model first and then use it to transcribe vast amounts of data to train a powerful E2E model~\cite{silovsky2023cross}.

This paper demonstrates that hybrid models can continue to outperform end-to-end models in low-resource language settings. Specifically, we present an approach that combines hybrid models with features extracted from self-supervised models, subword n-gram and RNN language models.
Furthermore, we find that replacing grapheme acoustic units with byte pair encoding (BPE)~\cite{sennrich2016neural} subword units yields better performance in code-switched scenarios.
We benchmark our approach on Gaelic, a Celtic language with 69,700 speakers in Scotland~\cite{scottish_census_2022}.
Despite the low number of speakers, Gaelic is somewhat atypical of low-resource languages because it has a long textual history with a standardised orthography, and a century of media broadcasts. %
These text and speech resources were previously used to build Gaelic ASR models with the standard hybrid recipes~\cite{rasipuram2013grapheme, evans2022developing}.
However, with our carefully tuned approach leveraging recent advancements,
  we achieve 54\% better performance than the previous models.
Furthermore, our approach is also 32\% better than our best fine-tuned Whisper model.

\section{Method}

To train the best possible Gaelic ASR system with a limited amount of manually transcribed data,
  it is important to leverage all available text corpora, which might be more accessible than transcribed speech.
Hybrid models~\cite{bourlard2012connectionist} can effectively combine acoustic models trained on transcribed data with language models trained on all available text.
Therefore, we train an acoustic model on the limited transcribed speech
  and train a language model on all available Gaelic text.
Since the amount of Gaelic text is limited compared to well-resourced languages,
  we experiment with using subword language models,
  which have been shown to work better than word language models with limited text data~\cite{smit2021advances}.
We also train RNN language models for use in lattice rescoring~\cite{mikolov2010recurrent} to better utilise the text data.

The performance can be boosted further by using large pre-trained models.
In particular, we use hidden representations extracted with very large multilingual pre-trained models~\cite{babu2022xlsr,chen2024towards} instead of MFCC features.
These large pre-trained models have seen very limited Gaelic data during pre-training.
Therefore, we run continued self-supervised pre-training on our training data~\cite{lee2022ctrl} to improve their performance on Gaelic data.

Languages like Gaelic with a strong focus on oral traditions may contain substantial quantities of speech data in media archives, 
and untranscribed speech might even be more readily available than text.
Therefore, it is important to be able to leverage this untranscribed speech: in this paper, we do it in two ways.
We use our initial ASR model to generate automatic transcriptions
  that are then manually corrected by professional transcribers,
  akin to active learning methods~\cite{drugman2016active}.
We also experiment with semi-supervised training~\cite{lamel2002},
  by directly using automatic transcriptions as pseudo-labels for training.

\section{Data}
\label{sec:data}

We trained our models on several datasets: see \tableref{tab:training_datasets}.
We used the same training data presented in \cite{evans2022developing} at the Celtic Language Technology Workshop, which we label ``CLTW train''.
This dataset consists of 103 hours of teaching videos, traditional narratives, and audio books,
  which were automatically aligned with normalised transcripts.
Upon analysing this data,
  we noticed that the transcripts were not normalised properly.
Therefore, to prevent any possible performance degradation,
  we applied our own normalisation script,
  which ensured that Gaelic accents are used consistently and which mapped words containing non-Gaelic letters to spoken noise.\footnote{We also noticed that apostrophes,
  which have a special role in Gaelic,
  were removed from the training transcripts.  It was not possible to fix this with normalisation, so
 we decided to ignore substitution errors with leading or trailing apostrophes during evaluation on this data.}

We also used 40 hours of manually transcribed speech from the historical BBC Radio nan Gàidheal programme, Prògram Choinnich, which ran for 30 years.
Finally, since code-switching between Gaelic and English is very common -- as in many marginalised languages --
  we also used 145 hours of accurately transcribed English broadcast data from the MGB dataset~\cite{bell2015mgb}.
In the later stages of our experiments we used an additional 68 hours of Prògram Choinnich data
  that was automatically transcribed with our initial ASR models and manually corrected by professional transcribers.
We also used 184 hours of untranscribed news data for semi-supervised training.

We used all available text data for training the language models.
This included manual transcripts, books and web data.
The most important datasets were the CLTW language model data (18M words), the Gaelic portion of MADLAD-400~\cite{kudugunta2024madlad} (86M words) and English text from MGB~\cite{bell2015mgb} (645M words).

We used four datasets for evaluating our model: see \tableref{tab:test_datasets}.
First, we used the test set from \cite{evans2022developing}, ``CLTW test'',
  which comprises 0.9 hours of speech.
Similar to the training data we found many problems with transcriptions of this dataset.
Therefore, our professional transcribers manually corrected the transcriptions.
The WER between original and corrected transcriptions was 16.3\%.
Due to its small size,
  we decided to use this dataset as a validation set during our experiments.
Since our ultimate goal is to provide subtitles for Gaelic broadcasts,
  we created two test sets from Gaelic BBC Alba programmes.
The first, ``News'', contains 7 episodes of an evening news programme broadcasted from 16th October 2023 to 22nd October 2023.
The second, ``BBC'', contains all Gaelic programmes broadcasted on 18th October 2023.
It consists of several genres ranging from kids cartoons, news, talk shows to drama.
Finally,
  we also held out 2 episodes of Prògram Choinnich (``PC'') as a test set.

\begin{table}
    \centering
    \caption{Summary of training datasets}
    \label{tab:training_datasets}
    \begin{tabularx}{0.9\columnwidth}{Xcc}
        \toprule
        \textbf{dataset}    & \textbf{language} & \textbf{\# hours} \\
        \midrule
        CLTW train~\cite{evans2022developing}      & Gaelic    & 103 \\
        Prògram Choinnich                          & Gaelic    & 40 \\
        MGB                                        & English   & 145 \\
        \midrule
        Extra Prògram Choinnich                    & Gaelic    & 68  \\
        News (untranscribed)                       & Gaelic    & 184 \\
        \bottomrule
    \end{tabularx}
    \vspace{1em}
    \caption{Summary of testing datasets}
    \label{tab:test_datasets}
    \begin{tabularx}{0.9\columnwidth}{Xc}
        \toprule
        \textbf{dataset}    & \textbf{\# hours} \\
        \midrule
        CLTW test~\cite{evans2022developing}          & 0.9 \\
        News                                          & 2.6 \\
        BBC                                           & 2.5 \\
        PC                                            & 1.7 \\
        \bottomrule
    \end{tabularx}
\end{table}

\section{Experiments}
\subsection{Whisper Fine-Tuning}
We fine-tuned Whisper~\cite{radford2023robust} with the HuggingFace Transformers library~\cite{wolf2020transformers} as a baseline approach for training Gaelic ASR.
Due to the limited amounts of transcribed training data (288 hours) and computation constraints,
  we decided to fine-tune Whisper-Turbo~\cite{radford2023robust} with LoRA~\cite{hu2022lora} using 8-bit quantisation of the frozen weights.
We trained the models on 4 NVIDIA GeForce RTX 2080 Titan GPUs for 20k steps,
  which corresponds to 11.3 epochs.
We used the default AdamW optimizer with 50 warm-up steps and a linear learning rate schedule with a peak learning rate 0.001. The effective batch size was 32.
We optimised the LoRA rank $r = \{64, 128, 256, 512\}$.

\subsection{Baseline Hybrid Models}

As a hybrid baseline,
  we used an improved version of the Gaelic ASR model from \cite{evans2022developing},
  which is available online via the Tar-sgrìobhadair API.
This model uses phoneme pronunciation lexicons and CNN-TDNN neural network architecture on top of MFCC features and i-vectors trained with LF-MMI~\cite{povey2016purely}.

We trained our own baseline hybrid models and all other subsequent acoustic models with the Kaldi toolkit~\cite{povey18_interspeech}.
Our baseline model was a standard TDNN-F model~\cite{povey18_interspeech} trained with LF-MMI~\cite{povey2016purely}.
It used MFCC features,
  but it did not use i\nobreakdash-vectors.
Even though a pronunciation lexicon exists for Gaelic,\footnote{\url{https://www.faclair.com/index.aspx}}
  we decided to use grapheme units to allow easier integration of subword n-gram language models.
We trained all n-gram language models with the SRILM toolkit~\cite{stolcke2002srilm}.
We experimented with a word 3-gram LM and a subword 4-gram LM trained on the training transcripts.
The subwords were produced by a BPE tokenizer~\cite{sennrich2016neural} with 10k tokens trained on the training transcripts.
Since the amount of training transcripts is limited,
  we also used other text sources as explained in \sectionref{sec:data}.

\subsection{SSL Features}

We used pre-trained SSL models as feature extractors to enhance the performance of our Gaelic ASR model.
We replaced traditional MFCC features with features extracted with SSL models as in \cite{lam2023comparing}.
In particular,
  we used two SSL models: XLS-R 300M~\cite{babu2022xlsr} and XEUS~\cite{chen2024towards}.
We extracted features from the 18th layer of both models.
Since neither of these models had been trained on Gaelic,
  we experimented with continued self-supervised pre-training of XLS-R 300M on our training data.
We used fairseq~\cite{ott2019fairseq} with the default XLS-R 300M pre-training configuration for additional 40k steps.
This took 2 days on 8 NVIDIA GeForce RTX 2080 Titan GPUs.
To be able to train on these GPUs with 12 GB VRAM,
  we reduced the maximum duration of each utterance to 5s.

Subsequently,
  we trained a standard TDNN-F hybrid model~\cite{povey18_interspeech} on top of these SSL features.
Since both SSL models output 50 features per second,
  we adapted the Kaldi training recipe to use a frame subsampling factor of 1 instead of the traditionally used 3.
We also removed the LDA initial layer,
  because estimating the transform is very slow for SSL features.

\subsection{BPE Acoustic Units}

In our initial experiments,
  we observed that our model did not perform well on code-switched utterances.
We hypothesized that this was due to the fact
  that graphemes are pronounced very differently in English and Gaelic.
Whilst this could be fixed by using phone pronunciation dictionaries instead of grapheme dictionaries, that solution would make using subword LMs complicated because it would be necessary to infer a pronunciation for each subword unit.
Instead,
  we decided to replace grapheme units with contextual graphemes as in~\cite{le2019senones, zhang2020faster}:
  we used BPE tokens~\cite{sennrich2016neural} as a replacement for graphemes,
  experimenting with various sizes of BPE inventory ($\{500, 1000, 2000, 5000\}$).
We used word-position independent BPE units,
  reducing the n-gram order of the denominator graph used for LF-MMI~\cite{povey2016purely} to 2 to prevent memory explosion.
Note that we still used `bi-phones' and we clustered them with tree-based clustering.

\subsection{RNN LM}

Our initial model made a lot of errors that we attributed to the weak language model.
Therefore, we applied RNN-LM rescoring~\cite{mikolov2010recurrent}.
We trained RNN-LMs with Kaldi on all available text training data mixed with English MGB text data for 40 epochs.
We trained three RNN-LMs with increasing size of the embedding and the LSTM hidden cells, respectively $\{(512, 128), (1024, 256), (2048, 512)\}$.

\subsection{Using Untranscribed Data}

To further improve the performance of our system,
  we investigated three approaches to increase the amount of training data.
First, we used noise augmentation~\cite{ko2017study} to make the models more noise robust,
  and to increase the amount of training data by using 3 noisy copies of each utterance.
Second, we used an earlier iteration of our ASR system to produce automatic transcriptions of 68 hours of the Prògram Choinnich data.
Human transcribers then manually corrected these automatic transcriptions.
The use of ASR made the manual transcription process $50\%$ faster compared to transcribing the recordings from scratch.
Third, we performed semi-supervised training~\cite{manohar2018semi} by using our ASR system to produce automatic transcription of 300~hours of An L\`{a} data.
We decoded this data in 30~s chunks
  and we segmented the data based on these first-pass transcripts.
We only used segments that were 5-30~s long for training, resulting in an additional 184 hours of training data.
Note that we applied noise augmentation to both Prògram Choinnich and An L\`{a} data and we pooled this data with the original noise augmented training data.

\begin{table*}[t]
    \centering
    \caption{Results}
    \label{tab:results}
    \footnotesize
    \begin{tabularx}{\textwidth}{ccccXccccc}
        \multicolumn{5}{l}{\textit{Baselines}} & \textbf{CLTW}    & \textbf{News}    &  \textbf{BBC} & \textbf{PC} & \textbf{Avg.}\\
        \midrule
        \multicolumn{5}{l}{Tar-sgrìobhadair API (an improved model from \cite{evans2022developing})}    & 20.2 & 23.7 & 38.7 & 29.2 & 28.0 \\
        \multicolumn{5}{l}{Whisper Turbo, LoRA 64}                                                      & 25.7 & 23.1 & 31.5 & 27.4 & 26.9 \\
        \multicolumn{5}{l}{Whisper Turbo, LoRA 128}                                                     & 21.9 & 21.0 & 28.4 & 22.8 & 23.5 \\
        \multicolumn{5}{l}{Whisper Turbo, LoRA 256}                                                     & 21.3 & 19.8 & 27.3 & 22.4 & 22.7 \\
        \multicolumn{5}{l}{Whisper Turbo, LoRA 512}                                                     & 20.8 & 19.2 & 26.9 & 21.2 & 22.0 \\
        \multicolumn{5}{l}{~~+ 68~hours of manually transcribed radio talk shows}                       & 19.5 & 17.1 & 25.8 & 19.3 & 20.4 \\
        \multicolumn{5}{l}{~~~~+ 184~hours of automatically transcribed news}                           & \textbf{19.4} & \textbf{15.0} & \textbf{23.5} & \textbf{18.1} & \textbf{19.0} \\      
        \midrule
        \\
        \multicolumn{5}{l}{\textit{Grapheme AM Baseline}} \\
        \midrule
        \textbf{AM Unit} & \textbf{AM Feature} & \textbf{LM Unit} & \textbf{LM Data} & \textbf{RNN LM} \\
        grapheme & MFCC & 300k words & train &           & 26.8 & 25.3	& 44.1 & 24.2 & 30.1 \\
        grapheme & MFCC & 10k BPEs  & train &  	      & 27.1 & 25.2	& 44.4 & 23.7 & 30.1 \\
        grapheme & MFCC & 10k BPEs  & web   &          & \textbf{26.5} & \textbf{21.7} & \textbf{42.8} & \textbf{22.6} & \textbf{28.4}\\
        \midrule
        \\
        \multicolumn{5}{l}{\textit{SSL Features}} \\
        \midrule
        grapheme & XLS-R 300M & 10k BPEs & web &          & 18.1 & 15.8 & 25.3 & 16.3 & 18.9 \\
        grapheme & XLS-R 300M CP & 10k BPEs & web &       & \textbf{16.0} & 15.9 & 24.2 & \textbf{14.7} & \textbf{17.7} \\
        grapheme & XEUS & 10k BPEs & web &            & 17.3 & \textbf{13.9} & \textbf{23.7} & 17.8 & 18.2 \\
        \midrule
        \\
        \multicolumn{5}{l}{\textit{BPE Acoustic Units}} \\
        \midrule
        0.5k BPEs & XLS-R 300M CP & 10k BPEs & web &	  & 14.9 & 13.9 & 22.1 & 14.3 & 16.3 \\
        1k BPEs & XLS-R 300M CP & 10k BPEs & web &	      & 14.6 & \textbf{13.7} & \textbf{21.6} & \textbf{14.0} & \textbf{16.0} \\
        2k BPEs & XLS-R 300M CP & 10k BPEs & web &	      & \textbf{14.4} & 14.0 & 21.7 & 14.1 & \textbf{16.0} \\
        5k BPEs & XLS-R 300M CP & 10k BPEs & web &	      & 15.0 & 14.7 & 22.2 & 14.2 & 16.5 \\
        \midrule
        \\
        \multicolumn{5}{l}{\textit{RNN LM Rescoring}} \\
        \midrule
        1k BPEs & XLS-R 300M CP & 10k BPEs & web & RNN LM 512  & 14.5 & 13.1 & 21.1 & 13.9 & 15.6 \\
        1k BPEs & XLS-R 300M CP & 10k BPEs & web & RNN LM 1024 & 13.9 & 12.7 & 20.5 & 13.6 & 15.2 \\
        1k BPEs & XLS-R 300M CP & 10k BPEs & web & RNN LM 2048 & \textbf{13.8} & \textbf{12.4} & \textbf{20.3} & \textbf{13.4} & \textbf{15.0} \\
        \midrule
        \\
        \multicolumn{5}{l}{\textit{Using Untranscribed Data}} \\
        \midrule
        \multicolumn{5}{l}{+ noise augmentation}                                               & 13.6 & 11.7 & 18.9 & 13.4 & 14.4 \\
        \multicolumn{5}{l}{~~+ 68~hours of manually transcribed radio talk shows}              & 13.1 & 11.2 & 18.0 & 12.5 & 13.7 \\
        \multicolumn{5}{l}{~~~~+ continual pre-training on all data}                           & \textbf{11.8} & 10.8 & 19.7 & \textbf{10.3} & 13.2 \\
        \multicolumn{5}{l}{~~~~~~+ 184~hours of automatically transcribed news}                 & 12.0 & \textbf{10.4} & \textbf{17.7} & 10.9 & \textbf{12.8} \\
        \bottomrule
    \end{tabularx}
\end{table*}

\section{Results}

We used the Tar-sgrìobhadair API and Whisper as baselines for our experiments.
The Tar-sgrìobhadair API achieved an average WER of 28.0\%.
Looking at fine-tuned Whisper-Turbo models,
  we see that the performance improves with the number of fine-tuned parameters.
The best average WER of 22.0\% was achieved with a LoRA rank 512.
This suggests that we might achieve even better results when fine-tuning the whole Whisper-Turbo model.
We also fine-tuned Whisper with the additional manually and automatically transcribed data, reducing the average WER to 19.0\%.
Our grapheme TDNN-F baseline model performed best with the web subword LM, with an average of WER 28.4\%, which is close to the Tar-sgrìobhadair baseline.

Subsequently,
  we explored the performance of models using SSL features instead of MFCC features.
We can see from the results that using SSL features makes a substantial difference,
  reducing the average WER by 33.5\% to 37.7\% relative.
XEUS appears to be a better base model than XLS-R 300M,
  but continued pre-training of XLS-R 300M, called XLS-R 300M CP, can achieve better performance than XEUS with an average WER of 17.7\%.
Whilst we believe that XEUS would also benefit from continued pre-training, unfortunately, code for this has not been integrated into ESPnet yet; therefore, we decided to use XLS-R 300M CP for the remainder of the experiments.

We next explored replacing graphemes with BPE units.
We can see that models using BPE units achieve 6.7\% -- 9.6\% lower WER than the model using grapheme units.
The best models use 1000 or 2000 BPE units and achieve an average WER of 16.0\%.
Therefore we decided to use 1000 BPE units.

After that, we rescored lattices with RNN language models.
Among these language models,
  the largest one achieves the best performance with an average WER of 15.0\%,
  a 6.3\% relative improvement compared to the first pass results.
We hypothesise that we could achieve even better performance by rescoring with a Transformer LM~\cite{vaswani2017attention} or a large language model fine-tuned on Gaelic.
In all the subsequent experiments, we report the WER obtained after rescoring with the largest RNN-LM.

Finally, we increased the amount of training data by noise augmentation, adding the additional manually transcribed data, and by using automatically transcribed data.
Furthermore, we make additional use of all the untranscribed data,
  we decided to continue pre-training XLS-R 300M on all our transcribed and untranscribed data for 100k iterations.
Combining all these techniques yielded the final average WER of 12.8\%,
  which is 32\% relative better than our best fine-tuned Whisper model.
Looking at individual test sets,
  we see that WER on CLTW, News and PC ranges from 10.4\% to 12.0\%.
However, the WER on the BBC test set is much higher at 17.7\%.
This is due to a large number of children's programs in the dataset,
  which have high WER because of deletion errors caused by background music and children’s voices.

Overall, these results show a promising trend, suggesting that we could achieve further gains by using larger amounts of untranscribed data.
We hope that we might further reduce the WER by doing continued self-supervised pre-training and lattice-based semi-supervised training with LF-MMI~\cite{manohar2018semi} on a much larger and more diverse corpus.

Since languages like Gaelic might have significantly more speech data available than text data,
  we also experimented with unsupervised language model adaptation~\cite{bacchiani2003unsupervised}.
We interpolated our web n-gram LM with an n-gram LM trained on automatic transcripts.
Our initial experiments showed that the adapted LM significantly degraded accuracy,
  therefore we left this for future experiments.

\section{Conclusions}

In this paper we showed that optimized hybrid models can achieve very good results on low-resource languages such as Scottish Gaelic.
Our model outperformed the previous best model deployed in the Tar-sgr\`{i}obhadair API by 54\%.
Furthermore, it outperformed our best fine-tuned Whisper-Turbo model by 32\% relative.
To get the best possible performance,
  we leveraged all available text data to train n-gram and RNN language models,
  used BPE acoustic units instead of graphemes,
  used SSL features instead of MFCCs,
  and performed noise augmentation, active and semi-supervised learning to increase the amount of training data.
We believe that this approach is applicable to all low-resource languages that have similar amounts of speech and text data available as Scottish Gaelic.

In the future,
  we will try fine-tuning other end-to-end models~\cite{pratap2024scaling,puvvada2024less} for Scottish Gaelic.
To get more transcribed data for training these end-to-end models,
  we plan to transcribe large quantities of untranscribed data with our best hybrid model as in \cite{silovsky2023cross}. 
Furthermore, based on the promising RNN LM results,
  we will attempt to fine-tune large language models on Gaelic,
  similar to recent work on Basque~\cite{etxaniz2024latxa}.
Such a model could be used for rescoring, generative error correction~\cite{chen2024hyporadise} or directly for speech recognition as in LLaMA-Omni~\cite{fang2025llamaomni}.

\section{Acknowledgements}

This work was supported by the Scottish Government (Grant name: ‘Ecosystem for Interactive Speech Technologies’). We thank BBC Alba for providing the data and Cailean Gordan, Alison Diack and Fearchar MacIllFhinnein for transcribing training and testing data.

\bibliographystyle{IEEEtran}
\bibliography{mybib}

\end{document}